\documentclass[letterpaper, 10 pt, conference]{ieeeconf}  

\usepackage{amsmath,amsfonts}
\usepackage{array}
\usepackage{algorithm}
\usepackage{algpseudocode}
\usepackage[caption=false,font=normalsize,labelfont=sf,textfont=sf]{subfig}
\usepackage{textcomp}
\usepackage{stfloats}
\usepackage{pifont}
\usepackage{hyperref}
\usepackage{booktabs} 
\hypersetup{
	colorlinks=true, 
	linkcolor=blue,  
	citecolor=blue, 
	urlcolor=red     
}
\usepackage{url}
\usepackage{verbatim}
\usepackage{graphicx}
\usepackage{xcolor}
\usepackage{tikz}
\usetikzlibrary{arrows.meta, positioning, fit, backgrounds, calc}
\def\BibTeX{{\rm B\kern-.05em{\sc i\kern-.025em b}\kern-.08em
		T\kern-.1667em\lower.7ex\hbox{E}\kern-.125emX}}
\usepackage{balance}

\IEEEoverridecommandlockouts        

\usepackage{amsmath}

\title{\Large \bf
SWIM: Vision-Language-Grounded Soft Whole-Body Interactive Manipulation}

\author{Tingcong Liu$^{1,4}$, Aye Phyu Phyu Aung$^{4}$, Junjie Xiong$^{3}$, Siyi Ma$^{3}$, Bo An$^{1}$, Ke Wu$^{3\dagger}$, Senthilnath Jayavelu$^{2,4\dagger}$\\
$^1$Nanyang Technological University, Singapore\\
$^2$National University of Singapore, Singapore \\
$^3$Mohamed bin Zayed University of Artificial Intelligence, Abu Dhabi, United Arab Emirates \\
$^4$Institute of Advanced Intelligence and Computing, A*STAR, Singapore \\ 
$\dagger$Corresponding Author
}

\begin{document}


\maketitle
\thispagestyle{empty}
\pagestyle{empty}

\begin{abstract}
Soft and continuum robots enable manipulation through distributed body deformation and contact, yet translating language and visual context into executable whole-body actuation remains a fundamental challenge. We present SWIM, a framework that maps an initial RGB observation and a language instruction to a complete actuation-command sequence.
Its vision-language-action (VLA) policy, SWIM-VLA, combines a diffusion
action head with Visual Soft Proprioception (VSP) through a shared
representation of RGB observations, language instructions, and tendon states.
The diffusion head models conditional distributions of expert
command chunks, while VSP supervises ordered body-anchor predictions
using simulation ground truth, encouraging the representation
to retain body geometry when learning from limited demonstrations.
Embodied mechanical intelligence supports physical execution of
command sequences generated through iterative virtual rollout
from evolving simulated observations, with intrinsic compliance
providing local contact adaptation without online policy queries.
We evaluate SWIM on packing, reaching, and grasping on a planar tendon-driven
soft robot, with grasping targets anchored.
In simulation, SWIM-VLA achieves success rates of 100\%, 96\%, and 88\%, respectively, outperforming an adapted OpenVLA-OFT baseline and controlled ablations.
On hardware, SWIM achieves success rates of 100\%, 80\%, and 75\%, compared with 75\%, 40\%, and 25\% for direct online deployment of the same policy checkpoint.
\end{abstract}



\section{Introduction}
\label{sec:introduction}

Soft and continuum robots combine structural compliance with large deformations, supporting applications in object handling, operation
in confined spaces, and minimally invasive
procedures~\cite{rus2015design,walker2013continuous}.
In object manipulation, their deformable bodies enable whole-body
wrapping and enclosure through distributed
contact~\cite{mcmahan2006field,li2016progressive}.
Research on this capability has pursued two complementary directions: regulating body shape and
configuration~\cite{pustina2023shape,almanzor2023staticshape},
and coordinating body motion and contact for
manipulation~\cite{marchese2014wholearm,graule2022contact}.

For configuration control, Pustina et al.~\cite{pustina2023shape} developed feedback laws to regulate prescribed soft-robot shapes under model uncertainty and constant disturbances, while Almanzor
et al.~\cite{almanzor2023staticshape} used deep visual inverse kinematics to reach user-specified
full-body target shapes.
Beyond configuration control, early OctArm experiments demonstrated teleoperated whole-arm grasping and object
handling~\cite{mcmahan2006field}.
Marchese et al.~\cite{marchese2014wholearm} planned whole-arm motion while accounting for the changing body envelope in confined environments.
Li and Xiao~\cite{li2016progressive} planned probing and progressive contact formation for whole-arm grasping in cluttered environments.
More recent work combined distributed tactile sensing with
hierarchical control for autonomous underwater
grasping~\cite{deldottore2026peripheral}, while Yang et al.~\cite{yang2026lightweight} learned whole-body grasping from actuation-space demonstrations.
These studies primarily address shape regulation and manipulation once a desired configuration or task-specific objective has been specified.
Instruction-driven whole-body manipulation introduces an additional task-grounding requirement: the robot must interpret the requested behavior and identify task-relevant targets from language and observed scenes.
This requires connecting semantic intent and visual context to physically executable whole-body actuation.

Vision-language-action (VLA) models map language instructions and visual observations to robot actions. Established formulations have largely been developed for rigid manipulators, using Cartesian end-effector motions, gripper states, or joint commands~\cite{zitkovich2023rt2,kim2024openvla,kim2025oft}. Recent work has extended vision-language-conditioned action
generation to soft and continuum robots.
These studies address end-effector-centered object
manipulation~\cite{su2025bridging,wei2026manisoft},
target tracking and navigation using discrete motor
commands~\cite{ng2025endovla,lin2026bilivla}, and the generation
of motion primitives and expressive whole-body behaviors through
low-level actuation~\cite{tang2026tmrvla,liu2026wholebody}. Extending these advances to manipulation through distributed body
contact requires addressing three mismatches.
First, an \emph{action-interface mismatch} arises because
end-effector motion does not fully specify the coordinated body
deformation and contact required for
manipulation~\cite{graule2022contact}.
A given task condition may also admit multiple valid actuation
sequences, motivating policies that capture conditional action
distributions~\cite{yang2026lightweight,chi2023diffusionpolicy}.
Second, a \emph{whole-body state mismatch} arises because
end-effector-centered state descriptions do not fully capture
the high-dimensional body configurations that govern soft
whole-body interactions~\cite{wei2026manisoft}.
Moreover, action targets provide only indirect supervision of body
geometry, making the relationship between body deformation and
actuation difficult to learn from limited demonstrations.
Third, an \emph{execution mismatch} arises from both sim-to-real
discrepancies and delays in online deployment.
For policies trained in simulation, visual differences can alter
policy inputs, while inaccuracies in robot and contact models
can change the physical response to predicted
commands~\cite{rao2021model,yang2026lightweight}. Inference and communication latency can further cause commands
to be applied after the body configuration and contact state have changed~\cite{kim2025oft,black2025realtime}.

Together, these mismatches complicate the translation of language
and visual context into whole-body manipulation, particularly when
demonstrations are limited and physical execution differs from
the conditions assumed during learning. Motivated by these challenges, we introduce SWIM, a framework for
vision-language-grounded soft whole-body interactive manipulation.
Our contributions are as follows:
\begin{figure*}[t]
   \centering
\includegraphics[page=6,width=170mm,trim=10mm 30mm 2mm 7mm,clip]{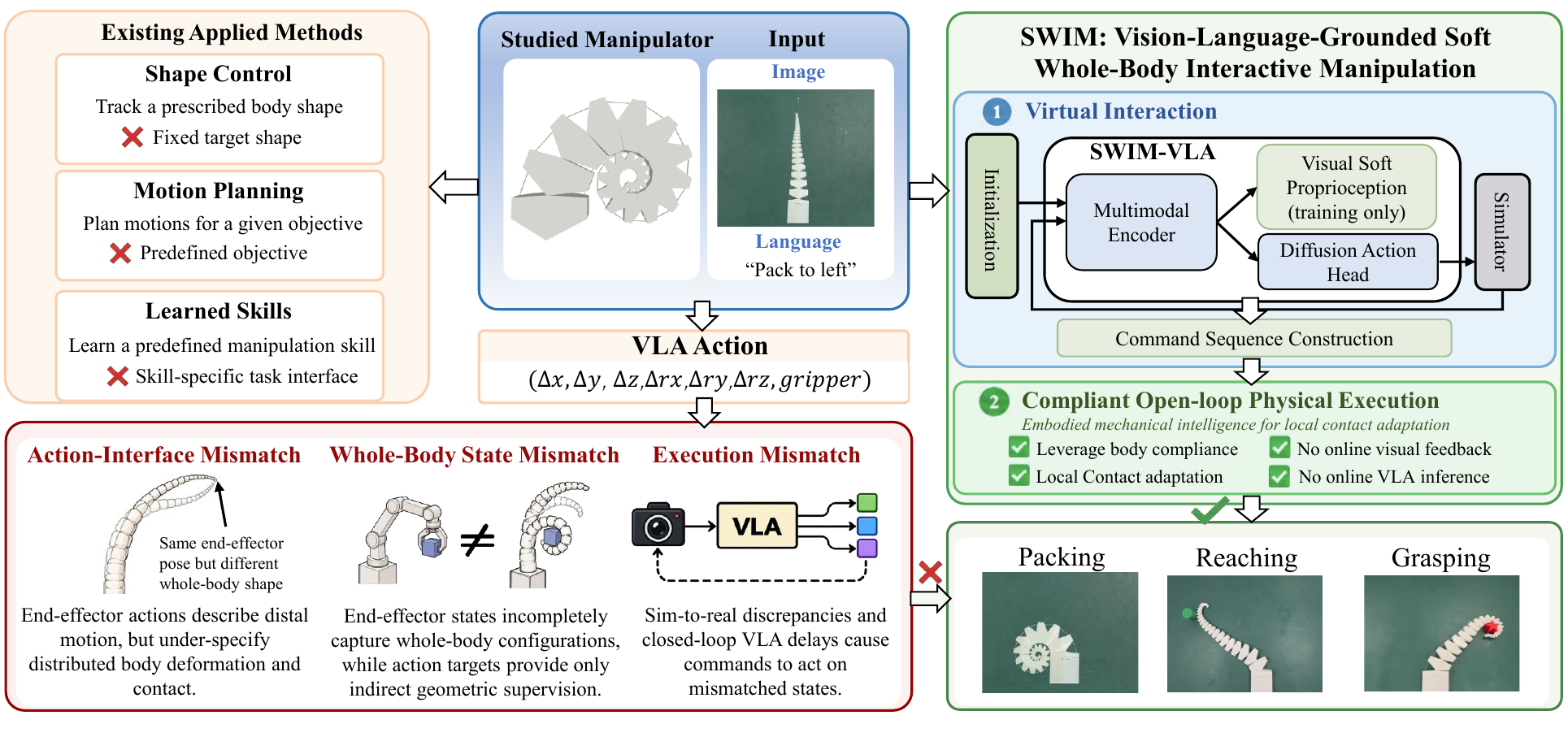}
   \caption{Motivation and design principles of SWIM for soft whole-body manipulation.}
   \label{fig:intro}
\end{figure*}
\begin{itemize}
\item \textbf{SWIM: virtual-rollout deployment supported by embodied mechanical intelligence.}
SWIM generates a complete actuation-command sequence from an initial
RGB observation and language instruction through virtual rollouts
using our proposed SWIM-VLA policy.
Intrinsic body compliance supports open-loop execution through local
contact adaptation, reducing centralized feedback demands while keeping
VLA inference outside the physical execution loop.
\item \textbf{SWIM-VLA: learning actuation distributions with geometric supervision.}
We introduce SWIM-VLA, a tendon-state-conditioned policy combining
a diffusion action head with Visual Soft Proprioception (VSP).
The diffusion action head models the conditional distribution of
expert command chunks, capturing alternative whole-body actuation
strategies under the same task conditions.
VSP uses ground-truth body-anchor coordinates from simulation to
encourage the shared representation to retain body geometry,
supporting learning from limited demonstrations.
    \item \textbf{Evaluation of policy design and physical deployment.}
   We evaluate packing, reaching, and grasping tasks on a planar tendon-driven robot, with targets anchored during grasping. Simulation ablations assess VSP and diffusion-based action prediction, while physical experiments compare virtual-rollout execution with direct online deployment of the same policy checkpoint.
\end{itemize}

\section{Problem Statement}
\label{sec:problem}

\subsection{Objective}
\label{subsec:objective}

We study vision-language-grounded whole-body manipulation for
soft robots.
A task is specified by an initial RGB observation $\mathbf{I}$
of the robot and task scene, together with a language
instruction $\ell$.
The desired output is a complete actuation-command sequence
\begin{equation}
\label{eq:task_output}
\mathbf{U}
=
\left[
\mathbf{a}_0,\ldots,\mathbf{a}_{T-1}
\right]^\top
\in\mathbb{R}^{T\times N_a},
\end{equation}
where $T$ is the number of command steps in the complete sequence,
$N_a$ is the number of independently controlled actuation channels,
and $\mathbf{a}_t\in\mathbb{R}^{N_a}$ is the command vector at step $t$.
We seek a sequence whose physical execution completes the
language-specified task through whole-body deformation and,
where required, distributed contact.

\subsection{The studied manipulator}
\label{sec:manipulator}
We evaluate the formulated problem on SpiRob, a tendon-driven
spiral continuum robot~\cite{wang2025spirobs}, as shown in Fig.~\ref{fig:intro}.
We use a planar SpiRob with a 24-joint logarithmic-spiral body
actuated by two independently driven tendons. Its monolithic
compliant structure enables large and continuous whole-body
deformation despite the low-dimensional actuation space.
Through the same actuation interface, SpiRob can produce diverse
behaviors, including extension, retraction, bending, curling,
wrapping, and enclosure, making it well suited for studying
coordinated whole-body deformation and distributed contact.

\section{Method}
\label{sec:method}

\begin{figure*}[t]
   \centering
\includegraphics[page=2,width=160mm,trim=5mm 0mm 10mm 0mm,clip]{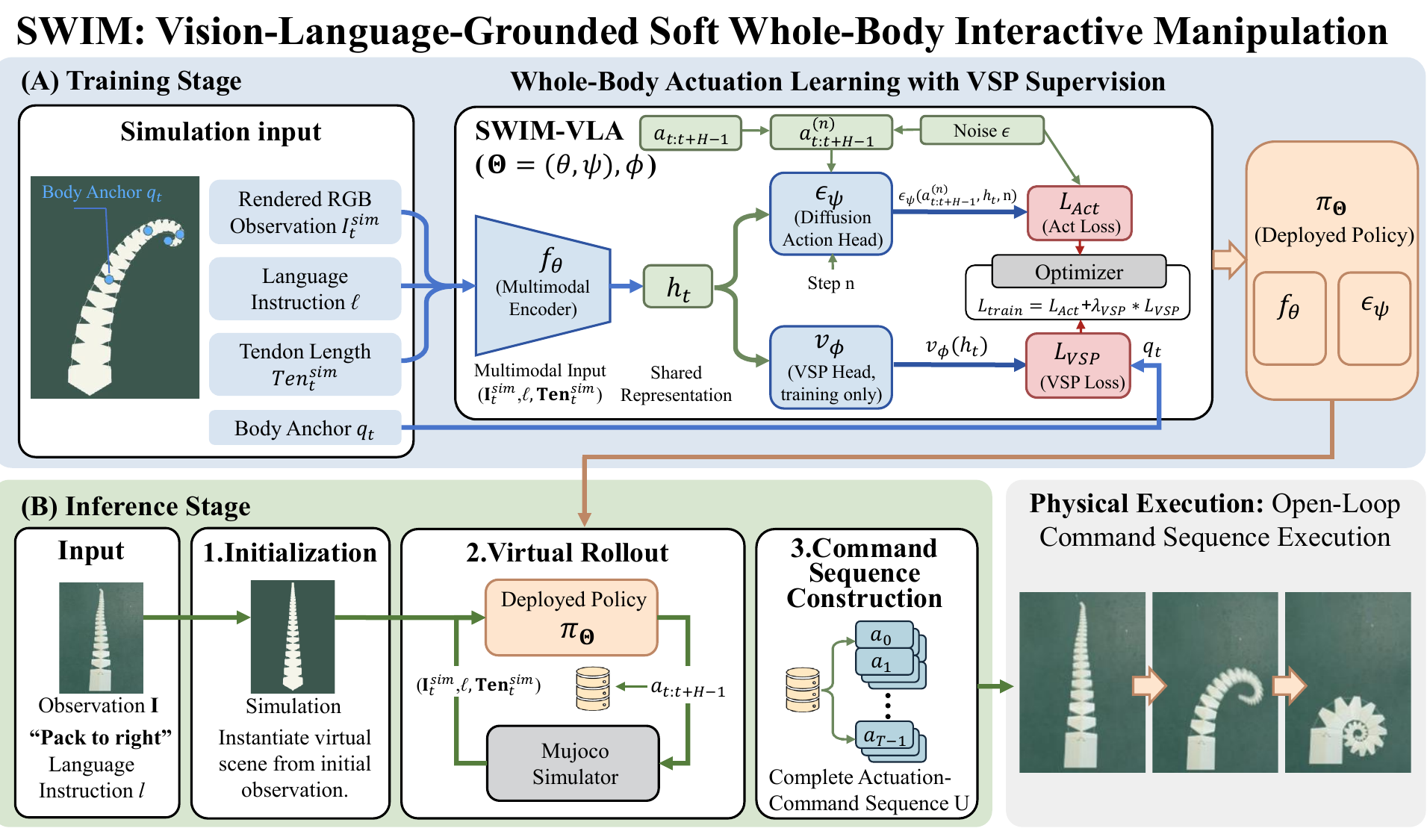}
   \caption{Overview of the SWIM framework. (A) SWIM-VLA is trained from simulated interaction trajectories using diffusion action learning with VSP supervision. (B) At inference, the trained policy performs virtual rollout to construct a complete actuation-command sequence, which is then executed open-loop on the physical robot.}
   \label{fig:framework}
\end{figure*}

SWIM maps an initial RGB observation $\mathbf{I}$ and language instruction
$\ell$ to a complete actuation-command sequence $\mathbf{U}$.
As illustrated in Fig.~\ref{fig:framework}, the framework consists
of two stages. During training, SWIM-VLA learns to predict command chunks through a diffusion action head
from simulated whole-body interaction trajectories with VSP supervision.
During inference, SWIM uses the trained policy to construct
$\mathbf{U}$ through virtual rollout, and the resulting
sequence is executed open-loop on the physical robot.

\subsection{SWIM-VLA Training from Simulated Demonstrations}
\label{subsec:training}

{SWIM-VLA builds on OpenVLA-OFT \cite{kim2025oft}, using the current tendon-length vector as a proprioceptive
input and representing policy actions as actuation commands.}
As shown in Fig. \ref{fig:framework}(A), SWIM-VLA is trained on simulated whole-body interaction
demonstrations. At simulated command step $t$, the policy predicts an
$H$-step actuation-command chunk from the multimodal input:
\begin{equation}
\label{eq:training_policy}
\begin{aligned}
\mathbf{a}_{t:t+H-1}
&\sim
\pi_\Theta
\left(
    \cdot
    \mid
    \mathbf{I}_t^{\mathrm{sim}},
    \ell,
    \mathbf{Ten}_t^{\mathrm{sim}}
\right).
\end{aligned}
\end{equation}
Here, $\mathbf{I}_t^{\mathrm{sim}}$ is the rendered RGB observation,
$\ell$ is the language instruction, and
$\mathbf{Ten}_t^{\mathrm{sim}}\in\mathbb{R}^{N_a}$ is the current
tendon-length vector. The tendon-length vector provides a
low-dimensional proprioceptive cue to the robot's current whole-body
configuration. The output
$\mathbf{a}_{t:t+H-1}\in\mathbb{R}^{H\times N_a}$ contains $H$
actuation commands for the $N_a$ tendon channels. We denote the deployable policy parameters by
$\Theta=(\theta,\psi)$.

\subsubsection{Shared Multimodal Representation}

The multimodal input is first encoded as
\begin{equation}
\label{eq:method_representation}
\mathbf{h}_t
=
f_\theta
\left(
   \mathbf{I}_t^{\mathrm{sim}},
    \ell,
    \mathbf{Ten}_t^{\mathrm{sim}}
\right),
\end{equation}
where $f_\theta$ is the multimodal encoder and $\mathbf{h}_t$ is the
shared multimodal representation. As depicted in Fig.~\ref{fig:framework}(A), $\mathbf{h}_t$ is passed to the diffusion action head $\epsilon_{\psi}$ for
command-chunk prediction and to the auxiliary VSP head $v_\phi$ for whole-body geometric
supervision.

\subsubsection{Actuation-Space Diffusion Objective}

{Prior work on the soft robot has shown that different sampled
actuation-command sequences can achieve the same whole-body
grasping task~\cite{yang2026lightweight}, suggesting that a task
condition may admit multiple feasible actuation realizations.
Motivated by this distributional structure, we adopt a diffusion
action head~\cite{chi2023diffusionpolicy,kim2025oft} to model the
conditional distribution over $H$-step command chunks spanning
all actuation channels.}

Given the shared multimodal representation $\mathbf{h}_t$,
a demonstrated command chunk is perturbed with Gaussian noise,
and $\epsilon_\psi$ is trained to predict the added noise:
\begin{equation}
\label{eq:action_loss}
\begin{aligned}
\mathbf{a}_{t:t+H-1}^{(n)}
&=
\sqrt{\bar{\alpha}_n}\,
\mathbf{a}_{t:t+H-1}
+
\sqrt{1-\bar{\alpha}_n}\,
\boldsymbol{\epsilon},
\\
\mathcal{L}_{\mathrm{act}}(\theta,\psi)
&=
\mathbb{E}_{t,n,\boldsymbol{\epsilon}}
\left[
    \left\|
        \boldsymbol{\epsilon}
        -
        \epsilon_\psi
        \left(
            \mathbf{a}_{t:t+H-1}^{(n)},
            n,
            \mathbf{h}_t
        \right)
    \right\|_2^2
\right].
\end{aligned}
\end{equation}
Here, $\theta$ and $\psi$ parameterize the multimodal encoder
$f_\theta$ and diffusion action head $\epsilon_\psi$, respectively.
$\mathbf{a}_{t:t+H-1}^{(n)}$ denotes the noisy command chunk at
diffusion step $n$, $\boldsymbol{\epsilon}\sim
\mathcal{N}(\mathbf{0},\mathbf{I})$ is the sampled Gaussian noise,
and $\bar{\alpha}_n$ is the cumulative noise-schedule coefficient
at step $n$.

\subsubsection{Visual Soft Proprioception}
End-effector states do not capture the full whole-body configuration,
while tendon lengths and action targets provide only indirect geometric
cues. We therefore introduce Visual Soft Proprioception (VSP), a
training-time auxiliary objective that supervises spatially
distributed body anchors and encourages $\mathbf{h}_t$ to retain
whole-body geometric information relevant to command prediction.

{As illustrated in Fig.~\ref{fig:framework} (A), let $\mathbf{q}_t \in [0,1]^{N_b\times 2}$ denote the normalized
image-plane coordinates of $N_b$ body anchors ordered along the
backbone from proximal to distal. Under the limited-demonstration
setting, these distributed anchors provide a compact geometric
descriptor covering both intermediate body locations and the
distal region. The VSP head $v_\phi$ predicts the anchor coordinates
using the following objective function:}
\begin{equation}
\mathcal{L}_{\mathrm{VSP}}(\theta,\phi)
=
\mathbb{E}_{t}
\left[
\frac{1}{N_b}
\sum_{j=1}^{N_b}
\left\|
[v_\phi(\mathbf{h}_t)]_j
-
\mathbf{q}_{t,j}
\right\|_2
\right].
\end{equation}
Here, $[v_\phi(\mathbf{h}_t)]_j$ and $\mathbf{q}_{t,j}$ denote
the predicted and ground-truth image-plane coordinates of the
$j$-th body anchor, respectively, and the loss averages the
Euclidean localization error over the $N_b$ ordered anchors.
By supervising multiple locations along the deformable body,
VSP encourages $\mathbf{h}_t$ to retain spatially distributed
geometric information useful for command-chunk prediction.

\subsubsection{Joint Training Objective}

The two objectives are jointly optimized as
\begin{equation}
\mathcal{L}_{\mathrm{train}}(\theta,\psi,\phi)
=
\mathcal{L}_{\mathrm{act}}(\theta,\psi)
+
\lambda_{\mathrm{VSP}}
\mathcal{L}_{\mathrm{VSP}}(\theta,\phi),
\end{equation}
where $\lambda_{\mathrm{VSP}}\geq 0$ controls the contribution of the
VSP supervision. During training, the multimodal encoder parameters
$\theta$ are optimized by both objectives, while $\psi$ and $\phi$ are
optimized by $\mathcal{L}_{\mathrm{act}}$ and
$\mathcal{L}_{\mathrm{VSP}}$, respectively. {After training, the auxiliary VSP head $v_\phi$ is discarded,
and $f_\theta$ and $\epsilon_\psi$ define the trained
SWIM-VLA policy $\pi_\Theta$.}

\subsection{Command-Sequence Construction via Virtual Rollout}

\label{subsec:inference}
As illustrated in Fig.~\ref{fig:framework}(B), SWIM constructs
a complete command sequence through iterative virtual rollouts
with the trained SWIM-VLA policy. The collected commands are
then executed open-loop on the physical robot, where intrinsic
body compliance supports local contact adaptation.

\subsubsection{Virtual Scene Initialization}

Given the initial RGB observation $\mathbf{I}$, SWIM initializes
a virtual scene using predefined object models and calibrated
planar locations~\cite{zhang2000flexible}. All objects are instantiated and included in the rendered
observation, allowing SWIM-VLA to predict target-directed
commands from the scene and language instruction $\ell$
during virtual rollout. The simulated robot starts from the
same predefined configuration used for physical reset.
The resulting initial policy input is
\begin{equation}
\mathrm{Init}(\mathbf{I},\ell)
=
\left(
\mathbf{I}^{\mathrm{sim}}_0,
\ell,
\mathbf{Ten}^{\mathrm{sim}}_0
\right),
\end{equation}
where $\mathbf{I}^{\mathrm{sim}}_0$ is the rendered RGB observation
of the initialized scene, and $\mathbf{Ten}^{\mathrm{sim}}_0$
is the tendon-length vector determined by the reset configuration.

\subsubsection{Virtual Rollout}

Starting from the initialized virtual scene, SWIM repeatedly
queries the trained SWIM-VLA policy defined in
Eq.~\eqref{eq:training_policy}. At each query, the policy predicts
an $H$-step actuation-command chunk, and the first $K$ commands
are executed in simulation:
\begin{equation}
\mathbf{a}^{\mathrm{exec}}_t
=
\mathbf{a}_{t:t+K-1},
\qquad K \leq H .
\end{equation}
The resulting RGB observation and tendon-length vector are then
used for the next policy query.

\subsubsection{Command Sequence Construction}

As shown in Fig.~\ref{fig:framework}(B), during virtual rollout, each executed command segment is appended
in temporal order to the command sequence:
\begin{equation}
\mathbf{U}
\leftarrow
\operatorname{concat}
\left(
\mathbf{U},
\mathbf{a}^{\mathrm{exec}}_t
\right).
\end{equation}
Repeating this process until task completion yields the complete
actuation-command sequence
\begin{equation}
\mathbf{U}
=
[\mathbf{a}_0,\ldots,\mathbf{a}_{T-1}]^\top ,
\end{equation}
where $T$ is the number of command steps in the complete
sequence. The sequence is then executed open-loop on the
physical robot without online queries to SWIM-VLA. This deployment scheme relies on embodied mechanical intelligence:
intrinsic body compliance accommodates local contact variations
during physical execution, while iterative VLA inference remains
within the virtual scene.

\section{Experimental Validation}
\label{sec:experiments}
{We evaluate the proposed framework from two aspects:
\textbf{(1) In simulation}, we evaluate SWIM-VLA on packing,
reaching, and grasping tasks, together with controlled ablations of VSP
and the diffusion action head; \textbf{(2) In physical deployment}, we evaluate SWIM by comparing
its virtual-rollout-based command-sequence execution with the direct
online execution of the same trained SWIM-VLA policy.}

\subsection{Experimental Setup and Evaluation Protocol}
\label{sec:preparation}
\subsubsection{Platform and Implementation}
\label{sec:exp_setup}
{Experiments are conducted using the SpiRob platform described in
Sec.~\ref{sec:manipulator}. We use matched robot geometry, tendon-actuation interfaces, and
top-down monocular observation configurations in MuJoCo simulation and on the
physical platform. SWIM-VLA uses OpenVLA-OFT as its backbone.
We jointly fine-tune a single policy on simulated demonstrations across different tasks. The principal
implementation and training settings are summarized in
Table~\ref{tab:implementation_details}.}
\begin{table}[t]
\centering
\caption{Implementation and training details of SWIM-VLA.}
\label{tab:implementation_details}
\footnotesize
\setlength{\tabcolsep}{3pt}
\renewcommand{\arraystretch}{1.08}
\begin{tabular}{p{0.28\columnwidth}p{0.65\columnwidth}}
\toprule
Setting & Value \\
\midrule

VLA backbone
& OpenVLA-OFT \cite{kim2025fine}. \\

Policy observation
& Top-down RGB image, language instruction, and current tendon-length vector. \\

Image resolution
& $256 \times 256$; OpenVLA-OFT image preprocessing. \\


Action space
& Absolute motor-torque commands
$\mathbf{a}_t\in\mathbb{R}^{2}$ for the two tendon drives,
using the same torque-space interface in simulation and
physical execution. \\

Action normalization
& Percentile-bounded normalization using the dataset $1$st and $99$th percentiles. \\

Command rate
& $10$ Hz. \\

Prediction horizon
& $H=15$ commands, corresponding to $1.5$ s. \\

Execution horizon
& $K=8$ commands (0.8 s) per policy query. \\

Diffusion training
& $50$ diffusion steps with a noise-prediction objective. \\

Diffusion inference
& $10$ DDIM denoising steps. \\

VSP targets
& $N_b=4$ ordered body anchors selected along the backbone at
approximately one-quarter, one-half, three-quarters, and the distal tip. \\

VSP head
& Two-layer MLP with a $512$-dimensional hidden layer, GELU activation,
and linear output. \\

VSP loss
& Mean Euclidean localization error over the $N_b$ ordered body anchors. \\

VSP loss weight
& $\lambda_{\mathrm{VSP}}=0.15$. \\

Fine-tuning
& LoRA with rank $32$ and bfloat16 mixed precision. \\

Optimizer
& AdamW. \\

Learning rate
& $3\times10^{-4}$ with cosine decay. \\

Batch size
& $4$ samples per GPU. \\

Training budget
& $20{,}000$ optimization steps. \\

Compute
& Four NVIDIA A100 GPUs. \\

Policy inference time & Approximately $700$\,ms per query (mean). \\

\bottomrule
\end{tabular}
\end{table}
\subsubsection{Tasks and Success Criteria}
Following the characteristic motion progression demonstrated by
SpiRob (packing, reaching, and grasping) \cite{wang2025spirobs}, we organize the
evaluation into three tasks of increasing interaction complexity. Given an initial RGB observation and a language instruction, SWIM must
perform the following tasks:

\begin{itemize}

\item \textbf{Packing.}
The instruction specifies a curling direction. The robot
must deform its body into a compact configuration in
the requested direction.

\item \textbf{Reaching.}
The instruction specifies a target object in the visual scene.
The robot must deform its body to bring its distal portion
close to the target.

\item \textbf{Grasping.}
The instruction specifies a target object in the visual
scene. The robot must approach and wrap its body around the target.
\end{itemize}
Task success is assessed from the final robot configuration
in each simulated rollout and physical test.
A packing trial is considered successful when the robot reaches
a fully curled configuration in the instructed direction.
A reaching trial is considered successful when the robot's
distal portion is close to the language-specified object
in the final configuration.
A grasping trial is considered successful when the robot wraps
its body around the language-specified target.

\subsubsection{Demonstration dataset}

We generate actuation-space demonstrations in MuJoCo following
the expert-demonstration procedure of prior whole-body grasping
work~\cite{yang2026lightweight} and the characteristic motion
patterns of SpiRob~\cite{wang2025spirobs}. Each training condition
defines a task setup, for which we collect four different successful
trajectories with distinct actuation-command sequences.
Each trajectory contains synchronized RGB observations, language
instructions, tendon lengths, and actuation commands, together with
ordered body-anchor coordinates for VSP supervision. For packing, a training condition is defined by the initial robot
configuration and the instructed curling direction. For reaching, the
instruction specifies a target object in the scene for the robot to
approach. For grasping, training conditions vary in target-object
position, color (red, green, and blue), and shape (cylinder,
pentagram, rectangular prism, and hexagonal prism). Reaching and
grasping scenes also contain visual distractors; only the
language-specified target participates in contact and dynamics.
During grasping, target objects remain rigidly anchored at their
sampled planar poses. Table~\ref{tab:simulation_dataset} summarizes the training
conditions and corresponding demonstration trajectories.

\subsection{Simulation Evaluation}
\label{sec:sim_eval}

We evaluate all methods on held-out test conditions not used
for training, including 50 packing, 50 reaching, and 100 grasping
conditions, with one rollout per condition.

\begin{table}[t]
    \centering
    \caption{\scriptsize Simulation demonstration and evaluation dataset.}
    \label{tab:simulation_dataset}
    \footnotesize
    \setlength{\tabcolsep}{3.5pt}
    \renewcommand{\arraystretch}{1.08}
    \begin{tabular}{lccc}
        \toprule
        Task & Train cond.  & Traj. per cond. & Train traj.\\
        \midrule
        Packing  & 18 & 4  & 72   \\
        Reaching & 32 & 4  & 128  \\
        Grasping & 148 & 4 & 592 \\
        \bottomrule
    \end{tabular}
\end{table}



\paragraph{Compared methods and ablations}
We compare the following complete policies and controlled SWIM-VLA
variants. All methods share the same backbone, inputs, data, and
training and evaluation settings.


\begin{itemize}

\item \textbf{OpenVLA-OFT.} We adopt OpenVLA-OFT~\cite{kim2025oft}, one of the VLA policies
evaluated in ManiSoft~\cite{wei2026manisoft}, and adapt it to the
same actuation-command space and multimodal inputs as SWIM-VLA.
The baseline retains its deterministic $L1$ action-prediction
objective.

\item \textbf{SWIM-VLA (Ours).}
The complete SWIM-VLA policy combines VSP supervision
with a conditional diffusion action head.

\item \textbf{SWIM-VLA without VSP.}
The auxiliary whole-body geometric objective is disabled by setting
$\lambda_{\mathrm{VSP}}=0$.

\item \textbf{SWIM-VLA without a diffusion action head.}
The conditional diffusion action head is replaced with a deterministic
$H$-step regressor trained using an $L_1$ objective.

\end{itemize}




\begin{figure}[!t]
   \centering
\includegraphics[page=3,width=95mm,trim=0mm 30mm 130mm 0mm,clip]{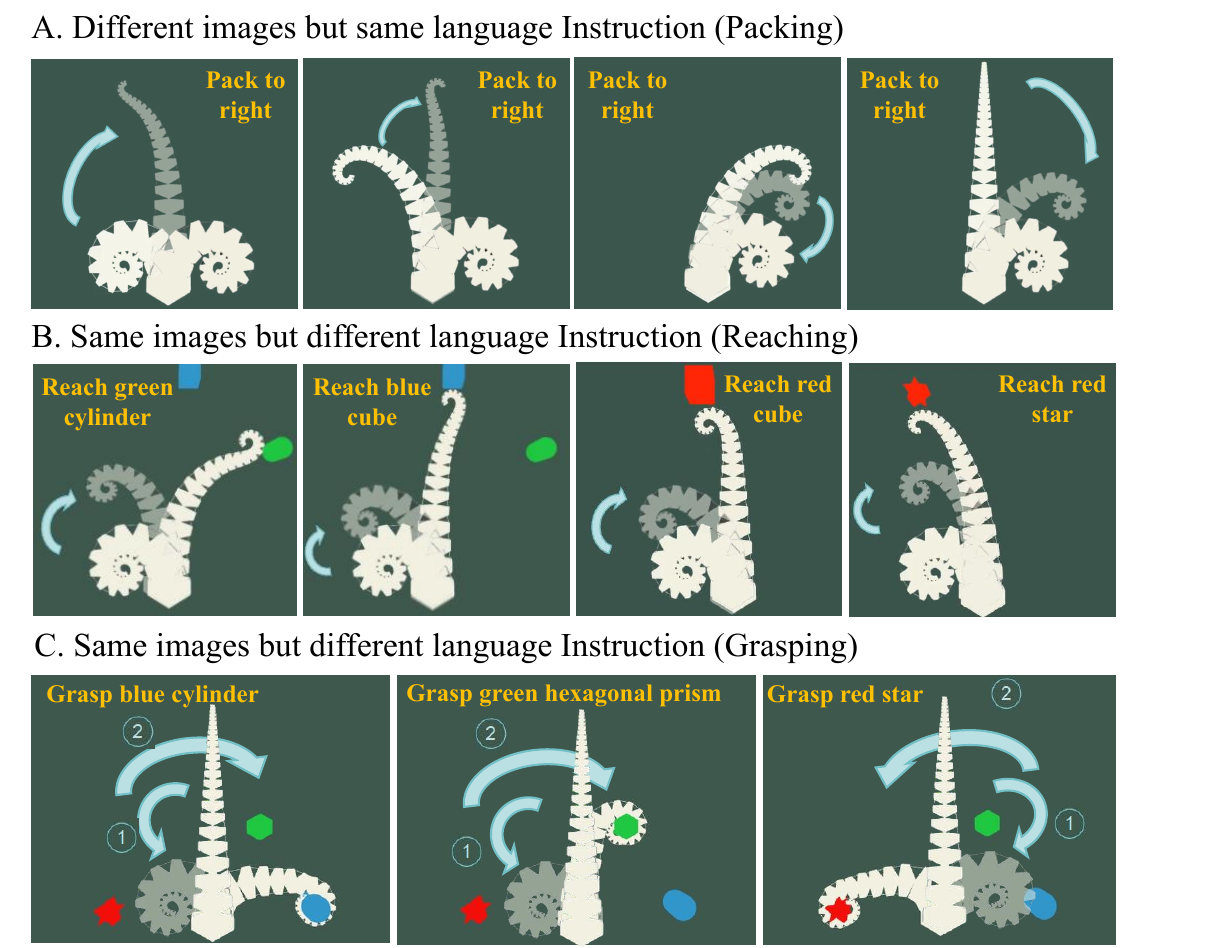}
   \caption{Qualitative evaluation of SWIM-VLA.}
   \label{fig:simulation_rollouts}
\end{figure}

\begin{figure*}[t]
   \centering
\includegraphics[page=1,width=180mm,trim=60mm 0mm 60mm 0mm,clip]{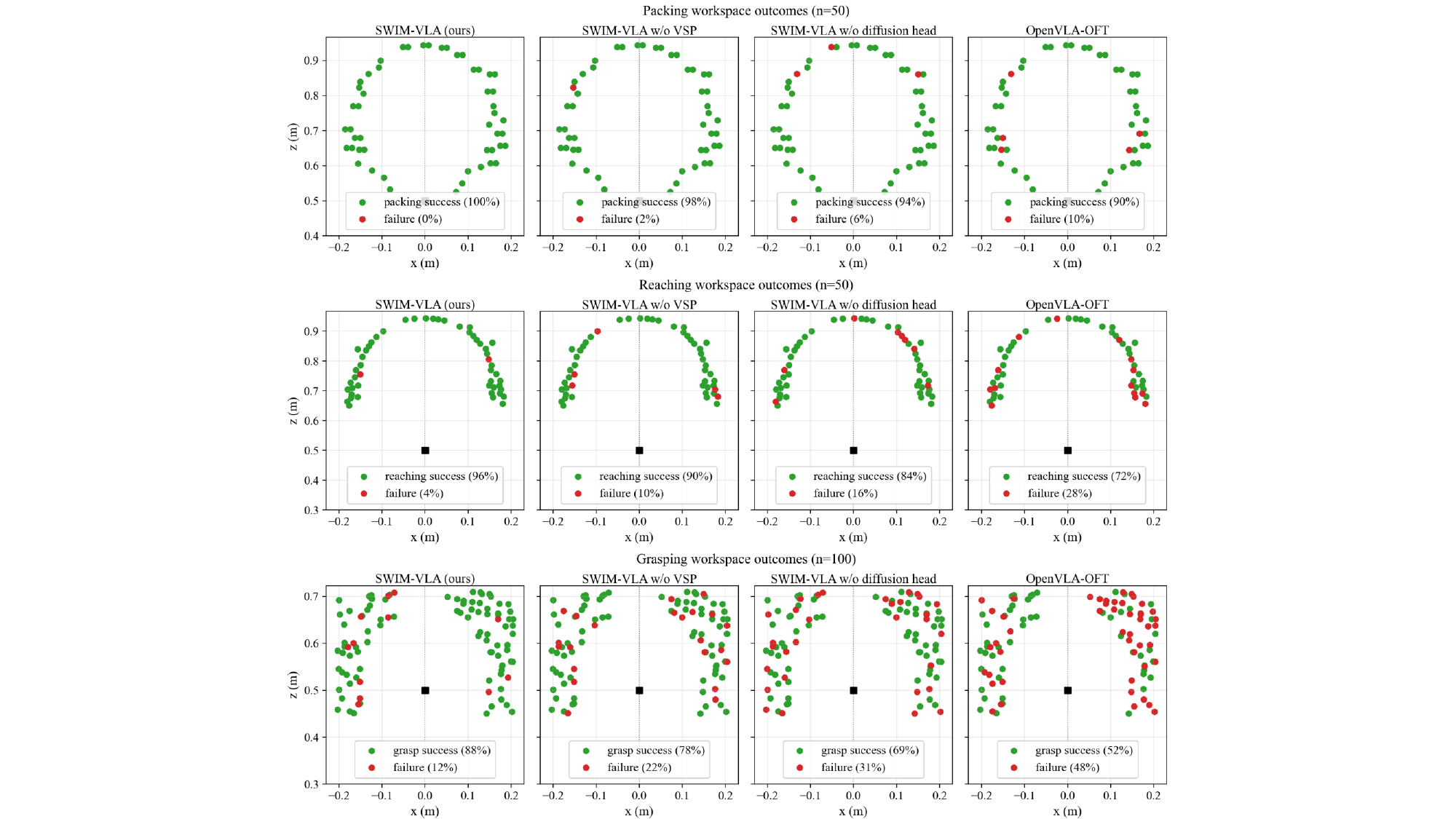}
   \caption{Spatial distributions of simulation outcomes.
Marker locations represent initial positions for packing
and target-object positions for reaching and grasping. Green and red markers indicate successful and failed trials,
respectively.}
   \label{fig:spatial_outcomes}
\end{figure*}

\begin{table}[h]
    \centering
    \caption{SIMULATION SUCCESS RATES (\%) ON HELD-OUT TEST CONDITIONS.}
    \label{tab:sim_results}
    \footnotesize
    \setlength{\tabcolsep}{3.5pt}
    \renewcommand{\arraystretch}{1.08}
    \begin{tabular}{lccc}
        \toprule
        Method & Packing & Reaching & Grasping \\
        \midrule

        OpenVLA-OFT
        & 90 & 72 & 52 \\

        \textbf{SWIM-VLA(Ours)}
        & \textbf{100} & \textbf{96} & \textbf{88} \\

        \midrule

        w/o VSP
        & 98 & 90 & 78 \\

        w/o diffusion action head
        & 94 & 84 & 69 \\

        \bottomrule
    \end{tabular}
\end{table}

\paragraph{Simulation results}
Table~\ref{tab:sim_results} reports success rates of $100\%$,
$96\%$, and $88\%$ for SWIM-VLA on packing, reaching,
and grasping, respectively, compared with $90\%$, $72\%$,
and $52\%$ for OpenVLA-OFT. The larger improvements between the two methods on reaching
and grasping suggest particular benefits for target-directed
deformation and external contact coordination. Fig.~\ref{fig:simulation_rollouts}(A) shows different motions
from different initial configurations under the same packing
instruction, illustrating the realization of a shared goal from
different states. Panel (B) shows the robot approaching the
language-specified object, while panel (C) shows that changing
the instruction within the same initial scene redirects target
selection and wrapping. Together, these examples show how
a jointly trained policy adapts whole-body behavior to the
current configuration and instructed goal. Fig.~\ref{fig:spatial_outcomes} further reveals the spatial
distribution of task outcomes. Packing remains robust across
the sampled initial positions, whereas reaching and grasping
show greater sensitivity to target location. Failures become
more frequent toward peripheral target regions when VSP or
diffusion is removed and are most widespread for OpenVLA-OFT.
In contrast, SWIM-VLA maintains more consistent success across
the evaluated workspace. Removing VSP lowers reaching and grasping success rates to $90\%$
and $78\%$, while packing remains at $98\%$
(Table~\ref{tab:sim_results}), supporting the value of geometric supervision
particularly for distal placement and contact formation.
Replacing diffusion with deterministic $L1$ regression
reduces grasping success from $88\%$ to $69\%$, with smaller
degradations in packing and reaching. Since each training
condition contains multiple successful trajectories with distinct
actuation-command sequences (Table~\ref{tab:simulation_dataset}),
the diffusion head is better suited to modeling the corresponding
conditional action distribution than a single deterministic
prediction. The larger degradation in grasping further suggests
that contact-rich whole-body interaction is more sensitive to
the quality of the predicted actuation sequence.

\subsection{Physical Evaluation}
\label{sec:physical_eval}

We validate the complete SWIM framework on the physical robot
using the trained SWIM-VLA policy. For each task, we conduct
20 physical trials for each deployment strategy.
\begin{figure}[b]
   \centering
\includegraphics[page=1,width=85mm,trim=0mm 20mm 0mm 30mm,clip]{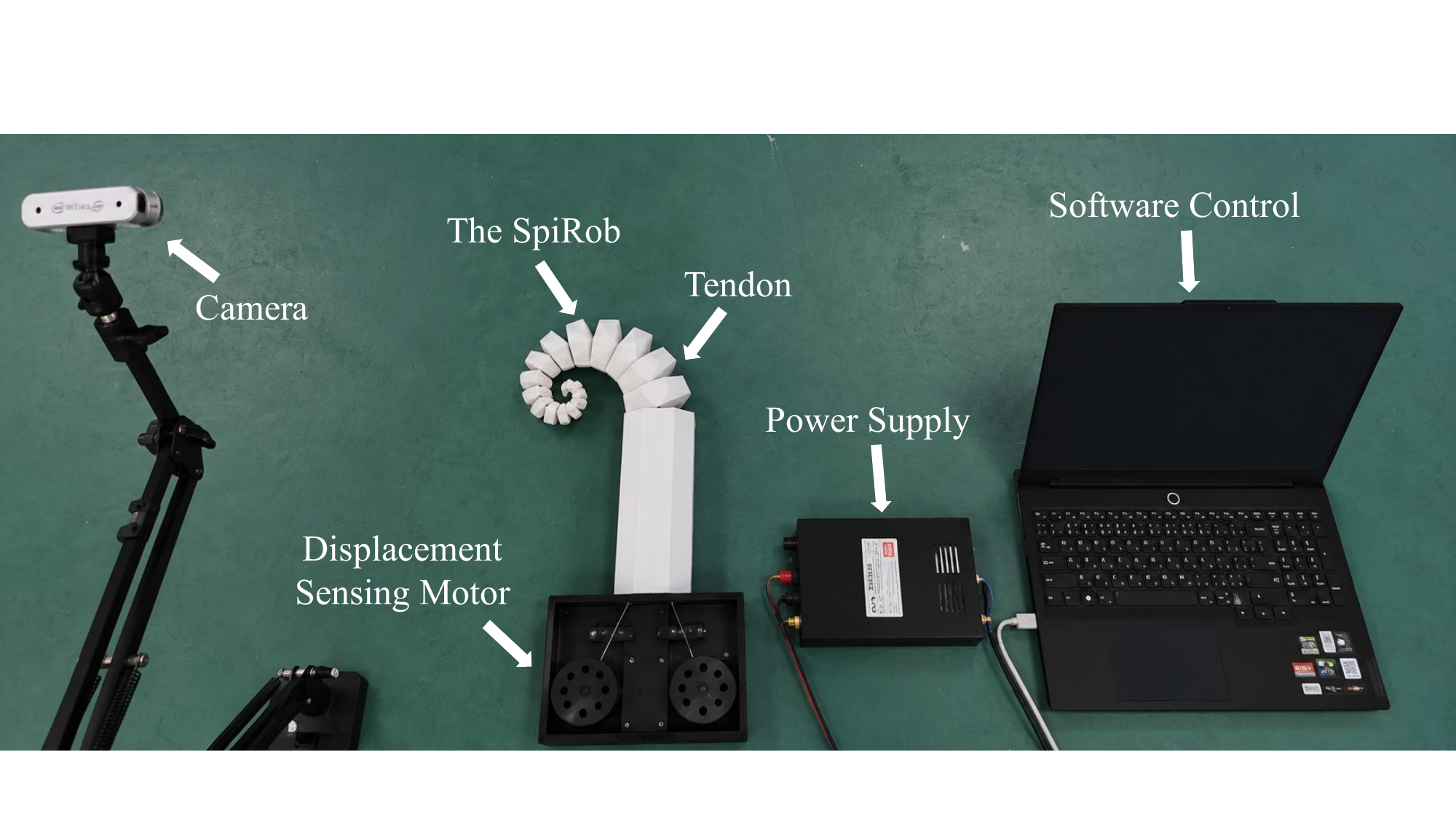}
   \caption{Physical deployment}
   \label{fig:physical_robot}
\end{figure}
The studied robot (Fig. \ref{fig:physical_robot}) operates in a planar workspace
observed by a fixed overhead RGB camera. Additional non-target distractors remain visible in the camera view but
do not participate in physical interactions.



\paragraph{Compared deployment strategies}
We compare two deployment strategies using the same trained SWIM-VLA
checkpoint.

\begin{itemize}

\item \textbf{Direct SWIM-VLA.}
The policy is queried online using physical RGB observations
and tendon-length vectors. At each query, it predicts a
15-step command chunk, of which the first 8 commands
are executed before the next policy query.

\item \textbf{SWIM.}
The initial physical observation is first used to initialize the virtual
scene. Task objects are segmented in HSV space, and their image centroids
are mapped to the MuJoCo workspace using a one-time planar
calibration~\cite{zhang2000flexible}. Predefined object models with
measured dimensions and matching colors are then placed at the estimated
planar locations. The same SWIM-VLA policy performs iterative rollouts
with the evolving simulated state, and the executed commands are collected
into a complete actuation-command sequence. The resulting sequence is
then executed open-loop on the physical robot without online policy
inference or replanning.

\end{itemize}


\begin{figure}[t]
   \centering \includegraphics[page=1,width=85mm,trim=0mm 50mm 130mm 0mm,clip]{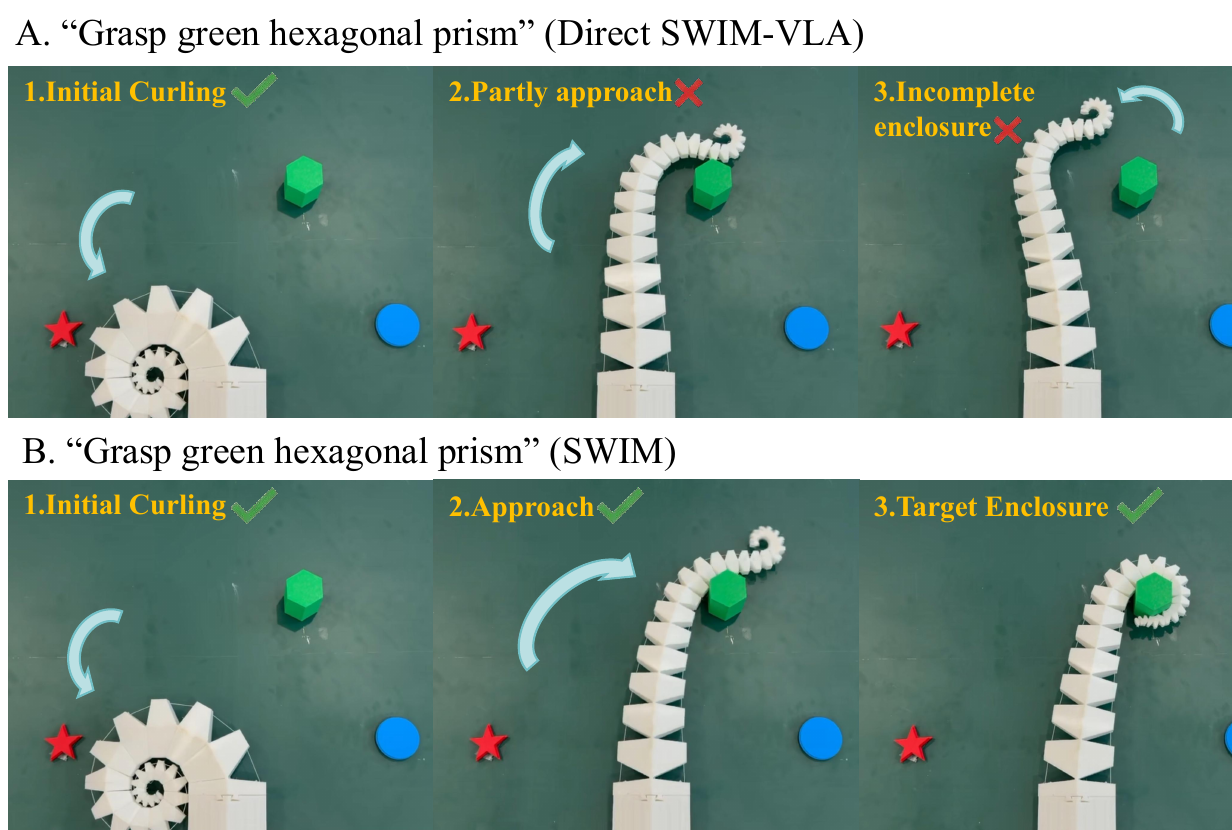}
   \caption{Physical grasping under the instruction
``Grasp green hexagonal prism'' using the same trained
SWIM-VLA checkpoint.
(A) Direct online deployment.
(B) SWIM with commands constructed by virtual rollout.}
   \label{fig:physical_comparison}
\end{figure}

\begin{table}[h]
    \centering
    \caption{Physical task success counts for direct SWIM-VLA
    deployment and the complete SWIM framework.}
    \label{tab:physical_comparison}
    \footnotesize
    \setlength{\tabcolsep}{4.5pt}
    \renewcommand{\arraystretch}{1.08}
    \begin{tabular}{lccc}
        \toprule
        Method & Packing & Reaching & Grasping \\
        \midrule
        Direct SWIM-VLA
        & 15/20 & 8/20 & 5/20 \\
        \textbf{SWIM}
        & \textbf{20/20} & \textbf{16/20} & \textbf{15/20} \\
        \bottomrule
    \end{tabular}
\end{table}

\paragraph{Physical deployment results}
SWIM achieves success rates of 100\%, 80\%, and 75\%
on packing, reaching, and grasping, respectively
(Table~\ref{tab:physical_comparison}), substantially
outperforming the direct deployment of the same SWIM-VLA
checkpoint, which achieves 75\%, 40\%, and 25\%. For reference, SWIM-VLA achieves success rates of 100\%, 96\%, and 88\%
in simulation (Table~\ref{tab:sim_results}). Thus, SWIM
recovers a substantial portion of the performance lost under
direct physical deployment, although reaching and grasping
remain below their simulation results. Fig.~\ref{fig:physical_comparison} illustrates incomplete
enclosure under direct deployment but successful wrapping
with SWIM. The degradation under direct deployment is
consistent with sim-to-real mismatch and online inference
latency (Table~\ref{tab:implementation_details}). Motivated by the robot's embodied mechanical intelligence,
SWIM shifts policy interaction to the virtual scene and executes
the resulting command sequence open-loop on the physical robot.
Intrinsic body compliance supports local contact adaptation,
reducing the need for repeated real-image feedback and online
VLA inference during execution. The remaining gap relative to simulation may arise from object-position
errors during scene initialization and differences in the robot's
material and structural properties, which can cause the same
commands to produce different physical deformations.

\section{Conclusion}
\label{sec:conclusion}

In this paper, we present SWIM, a framework for vision-language-grounded
whole-body manipulation with soft robots. SWIM-VLA combines
actuation-space diffusion prediction with Visual Soft
Proprioception to learn language-conditioned whole-body
commands from limited simulated demonstrations. At deployment,
SWIM constructs a complete command sequence through virtual
rollouts and executes it open-loop on the physical robot, where
body compliance supports local contact adaptation. Simulation results show the benefits of both VSP and diffusion-based
action prediction, while physical experiments demonstrate that SWIM
substantially outperforms the direct online deployment of the same
SWIM-VLA checkpoint and recovers much of the performance lost
during direct sim-to-real transfer. The current study is limited to
planar manipulation, predefined reset configurations, and anchored
grasping targets. Future work will consider more general 3D
interactions, moving objects, and improved scene and physical-model
alignment.


\bibliographystyle{IEEEtran}
\bibliography{reference}

\end{document}